# SV3.3B: *A Sports Video Understanding Model for Action Recognition*

Sai Varun Kodathala
*Research and Development*
*Sports Vision, Inc.*
Minnetonka, USA
varun@sportsvision.ai

Yashwanth Reddy Vutukoori
*Research and Development*
*Sports Vision, Inc.*
Minnetonka, USA
yashwanth@sportsvision.ai

Rakesh Vunnam
*Research and Development*
*Vizworld, Inc.*
Minnetonka, USA
rakesh@vizworld.ai

*Abstract*— This paper addresses the challenge of automated sports video analysis, which has traditionally been limited by computationally intensive models requiring server-side processing and lacking fine-grained understanding of athletic movements. Current approaches struggle to capture the nuanced biomechanical transitions essential for meaningful sports analysis, often missing critical phases like preparation, execution, and follow-through that occur within seconds. To address these limitations, we introduce SV3.3B, a lightweight 3.3B parameter video understanding model that combines novel temporal motion difference sampling with self-supervised learning for efficient on-device deployment. Our approach employs a DWT-VGG16-LDA based keyframe extraction mechanism that intelligently identifies the 16 most representative frames from sports sequences, followed by a V-DWT-JEPA2 encoder pretrained through mask-denoising objectives and an LLM decoder fine-tuned for sports action description generation. Evaluated on a subset of the NSVA basketball dataset, SV3.3B achieves superior performance across both traditional text generation metrics and sports-specific evaluation criteria, outperforming larger closed-source models including GPT-4o variants while maintaining significantly lower computational requirements. Our model demonstrates exceptional capability in generating technically detailed and analytically rich sports descriptions, achieving 29.2% improvement over GPT-4o in ground truth validation metrics, with substantial improvements in information density, action complexity, and measurement precision metrics essential for comprehensive athletic analysis.

## I. Introduction

Over the past two decades, the paradigm of sports performance management has evolved from traditional, isolated evaluation and conditioning to a more holistic, function-oriented approach. This progression emphasizes integrating visual analysis into sports training. This perspective shift underscores the belief among practitioners that a comprehensive understanding of sports action sequences is crucial. Such knowledge empowers athletes to refine their skills, coaches to provide more practical guidance, and performance analysts to identify and rectify performance deficits.

Professional video analytics tools and expertise have historically been concentrated in elite sports organizations with substantial budgets, sophisticated multi-camera setups, and dedicated analysis teams. This has created a significant divide, where most amateur athletes and coaches worldwide lack access to the visual insights that could dramatically improve their development. Visual understanding of sports videos helps us: (1) democratize access to professional-grade analytics for resource-constrained environments prevalent in amateur sports, addressing critical market needs where 70% of coaches cannot effectively recall or analyze performance [1]. (2) transform basic mobile footage into comprehensive multimodal insights that connect coaching notes with video segments, tactical diagrams with game footage, and enable natural language querying of video content. (3) capture fine-grained motion details while simultaneously identifying broader tactical patterns. (4) provide meaningful analysis despite amateur sports' unique challenges including varying camera angles, limited available shots, and inconsistent lighting conditions. (5) address critical safety and development needs in youth and amateur sports where injury rates exceed professional leagues due to limited access to proper analysis tools.

Unlike static videos, sports videos contain complex motion, spatial, and temporal features vital for understanding multi-phase actions. This necessitates using cross-modal features to grasp the temporal understanding of sports videos. While contemporary Multimodal Large Language Models (MLLMs) have demonstrated significant progress in building perception agents capable of understanding visual sequences, most of them are computationally intensive and require server-side processing, leading to extremely high costs for sports applications. The requirements for on-device deployment in sports environments, considering connectivity limitations during training sessions, real-time analysis latency constraints, and privacy considerations for athlete data, demand lightweight solutions that preserve comparable accuracy levels. Although model compression efforts are ongoing, current approaches remain substantially larger than the maximum model size that can run efficiently on mobile devices while maintaining acceptable inference speeds.

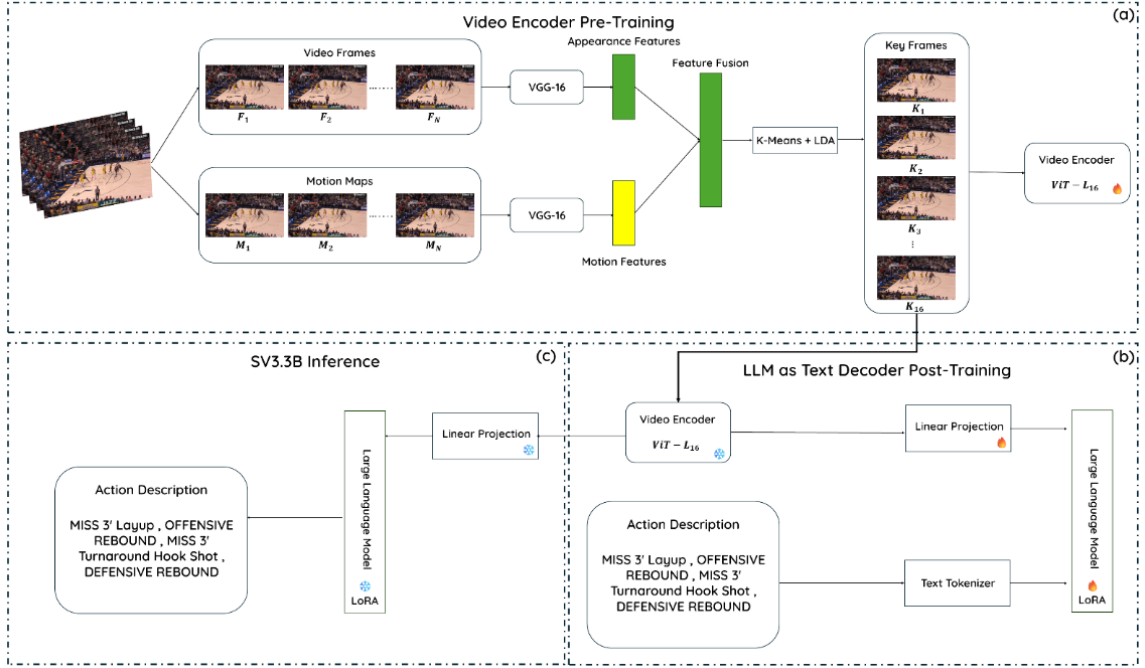

Fig. 1. **SV3.3B Architecture Overview.** The framework operates through three main phases: (a) Video Encoder Pre-Training employs DWT-VGG16-LDA sampling to extract 16 keyframes from input video sequences, processes appearance and motion features through dual VGG-16 pathways, applies K-means clustering and LDA for feature fusion, and trains a ViT-L encoder using JEPA2 self-supervised learning; (b) LLM as Text Decoder Post-Training fine-tunes a LLaMA-3.2-3B model with LoRA adaptation, using a linear projection layer to map frozen video encoder features to the LLM embedding space; (c) SV3.3B Inference combines the pretrained video encoder with the fine-tuned LLM decoder through linear projection to generate sports action descriptions from input videos.

Previous methods like Video-ChatGPT [2] use spatial and temporal pooling that loses critical information, while LLama-VID's extreme compression to two tokens per frame cannot capture essential biomechanical transitions [3]. These systems can identify broad events (e.g., "free throw") but lack granular understanding. Meaningful sports analysis requires detection of distinct phases like preparation (positioning), launch (initiating motion), release (execution mechanics), and follow-through, all occurring within seconds. These transitions contain vital information about technique and performance that existing methods fail to preserve.

To address these limitations and bridge the gap between computational efficiency and detailed sports understanding, we draw inspiration from self-supervised learning approaches, particularly Joint Embedding Predictive Architectures (JEPA) [4] and their variants [5, 6]. Adapting the successful framework from UI-JEPA [7], we introduce V-DWT-JEPA2 (SV3.3B), a lightweight video-to-text model specifically designed for sports activity analysis. Our approach employs a JEPA2 - based encoder [8] that leverages novel temporal motion difference sampling techniques on unlabeled sports video data to extract abstract feature representations. Following the UI-JEPA paradigm, these representations are then processed by an LLM decoder that generates comprehensive sports action descriptions. Unlike UI activities that focus on user intent prediction, our model is tailored to capture the nuanced biomechanical transitions and multi-phase action sequences inherent in sports movements, making it particularly suitable for fine-grained sports analysis where temporal precision is critical.

## II. RELATED WORK

The broader landscape of sports understanding [9] encompasses diverse research domains including action spotting [10, 11, 12], commentary generation [13, 14, 15, 16, 17], athlete performance analysis [18, 19], tactical planning [20], and intelligent referring systems [21, 22]. Recent developments in multimodal large language models have prompted researchers to create unified frameworks [23, 24, 25] capable of handling multiple sports understanding tasks simultaneously. While visual-language models [26, 27, 28, 29, 30] have achieved remarkable success in general applications, their extension to video understanding [31, 32, 33] presents additional complexities including temporal alignment [34, 35], dense captioning [36, 37, 38], and audio description [39-41]. However, their applicability to specialized professional domains like sports remains limited due to the unique temporal dynamics, domain-specific terminology, and nuanced movement patterns that characterize athletic activities, necessitating purpose-built solutions rather than generic adaptations.

Our work with V-DWT-JEPA2 addresses this fundamental challenge of extracting meaningful insights from sports videos while operating under severe computational constraints. We develop a model that automatically identifies and describes complex multi-phase athletic movements, from preparation through execution and follow-through, without requiring extensive labeled datasets or high-end hardware. Through our novel DWT-VGG16-LDA based sampling approach integrated with JEPA2 architecture, we achieve detailed sports action understanding that rivals computationally intensive MLLMs while maintaining the efficiency necessary for real-time analysis on mobile devices. This research bridges the gap between professional-grade sports analytics and accessible technology, demonstrating that sophisticated movement analysis can be democratized through efficient model design and targeted self-supervised learning strategies.

### III. THE V-DWT-JEPA2 FRAMEWORK

*A. Architecture Design:*

*1) Problem Formulation:*

Given an input sports video clip $X \in R^{H \times W \times 3 \times N}$ consisting of $N$ RGB frames of size $H \times W$, we aim to extract $K$ key frames (where $K << N$) and predict a sequence of descriptive text $\{y\}$ that accurately characterizes the sports action captured in the video. The sequence $\{y\}$ has arbitrary length and must capture the technical nuances of the action sequence. The challenge lies in efficiently determining the optimal value of $K$ and extracting the most relevant visual information from the high-dimensional input $X$ while preserving the temporal dynamics necessary for accurate action description.

*2) Overall Structure:*

The SV3.3B framework operates through a three-phase architecture as illustrated in Fig. 1. In Phase (a), the input video sequence undergoes robust temporal sampling through our DWT-VGG16-LDA mechanism, which reduces the N-frame sequence to K=16 representative keyframes that capture critical biomechanical transitions. These selected frames are processed through dual-path feature extraction, where appearance features are extracted using VGG-16 to capture static visual elements, while motion features are computed from wavelet approximation coefficient differences between consecutive frames to represent dynamic movement patterns. The resulting feature vectors undergo K-means clustering followed by Linear Discriminant Analysis for optimal fusion, creating a unified high-dimensional representation that simultaneously encodes spatial appearance and temporal dynamics. These keyframes are then used to train a Vision Transformer Large (ViT-L) encoder with 300M parameters using the JEPA2 self-supervised learning framework through mask-denoising objectives.

In Phase (b), the frozen pretrained video encoder is combined with a fine-tuned LLaMA-3.2-3B decoder employing LoRA adaptation. The 1024-dimensional spatiotemporal representations from the video encoder are processed by a linear projection module that maps features to the LLM's embedding space, enabling the generation of natural language descriptions of sports actions. Phase (c) demonstrates the complete SV3.3B inference pipeline, where input videos are processed through the integrated architecture to produce detailed sports action descriptions. This design enables the model to capture fine-grained biomechanical understanding while maintaining computational efficiency suitable for edge device deployment, with the complete system comprising 3.3B parameters optimized for sports-specific video analysis tasks.

*B. The DWT-VGG16-LDA based Sampling*

To improve temporal information extraction from sports video sequences for the V-DWT-JEPA2 framework, we implement a robust keyframe selection mechanism that identifies the most representative frames corresponding to critical phases in athletic movements. Rather than relying on uniform temporal sampling, which may miss crucial biomechanical transitions or include redundant information, this approach systematically processes video content to extract keyframes that provide a comprehensive yet concise visual summary of complex motion sequences.

The keyframe extraction algorithm builds upon the combination of Discrete Wavelet Transform (DWT) for motion analysis, VGG-based feature extraction for appearance understanding, and Linear Discriminant Analysis (LDA) for optimal feature fusion. The algorithm computes a wavelet approximation $(A_i, a_i)$ for each sampled frame $F_i$ using DWT at the second decomposition level ($L = 2$) with Haar wavelet. This specific configuration preserves fine-grained motion details while maintaining computational efficiency, making it particularly suitable for capturing rapid biomechanical transitions such as release mechanics in basketball shooting or quick directional changes in athletic movements. The Haar wavelet provides excellent localization properties in both time and frequency domains, enabling effective detection of abrupt changes in motion patterns that characterize critical phases in sport actions.

The algorithm implements a dual-path feature extraction approach, processing appearance and motion information through parallel streams. The original RGB frames are processed through the VGG-16 model to extract appearance features $(f_a, i)$ that capture static visual elements such as posture and position. Simultaneously, motion maps $(D_i, d_i)$ are calculated as the difference between consecutive wavelet approximation coefficients ($LL$) of consecutive frames rather than raw pixel differences, resulting in a more robust motion representation that remains stable under varying lighting conditions. These motion maps are converted to 3-channel images and processed through the same VGG model to extract motion features $(f_m, i)$ representing dynamic movement patterns. These complementary feature vectors are collected into separate arrays ($F_a$ and $F_m$) throughout the frame sampling process. The critical fusion step combines these feature arrays into a unified representation $F = [F_a, F_m]$, creating a high-dimensional feature space that simultaneously encodes spatial appearance and temporal dynamics.

The parameter $K$ in the algorithm specifies the number of keyframes to be extracted, which directly corresponds to the number of distinct action phases in a sports motion sequence. For the V-DWT-JEPA2 implementation, $K$ is set to 16 to align with the input requirements of the ViT-L JEPA2 architecture, ensuring optimal coverage of multi-phase athletic movements while maintaining computational efficiency. After feature fusion, $K$-means clustering is applied to the combined feature space $F$ to obtain pseudo-labels that identify natural groupings of similar frames. Linear Discriminant Analysis is then employed for dimensionality reduction of the fused features, maximizing the separation between the identified clusters while preserving their internal coherence. Within this LDA-transformed space, the algorithm computes the center of each cluster and selects the frame whose feature representation is closest to each center as a keyframe as shown in Fig. 2. This precise selection mechanism ensures that each keyframe represents a distinct phase of the action sequence while collectively providing comprehensive coverage of the entire motion, thereby enhancing the temporal understanding capabilities of the sports video analysis framework.

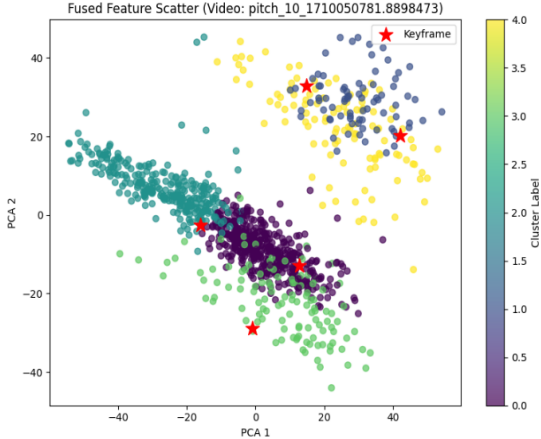

Fig. 2. PCA Projection of Fused Feature Space with Selected Keyframes.

The wavelet-VGG feature fusion method demonstrates superior performance compared to conventional sampling approaches when evaluated across multiple sports domains. As shown in experimental validation in Fig. 3, 4 and 5, this method significantly outperformed uniform sampling and color histogram methods in identifying key biomechanical phases, demonstrating robust performance across basketball free throws, golf swings, and other athletic movements beyond the initial baseball evaluation domain.

C. *Training Pipeline*

　a) *Video Encoder Pre-Training:*

The video encoder employs a self-supervised pretraining strategy using the Video - Joint Embedding Predictive Architecture 2 (V-JEPA2) framework with mask-denoising objectives. The encoder utilizes a Vision Transformer Large (ViT-L) architecture with 300 million parameters, processing video inputs as sequences of 16 frames sampled using the DWT-VGG16-LDA keyframe extraction method. The self-supervised learning approach trains the model to predict masked portions of video sequences in the learned representation space, enabling the model to learn meaningful spatiotemporal representations without requiring labeled data. The masking strategy applies to two different configurations: 8 blocks with spatial masking scale [0.15, 0.15] and 2 blocks with spatial scale [0.7, 0.7], both maintaining temporal scale [1.0, 1.0] and aspect ratios between [0.75, 1.5].

The training configuration in this work processes videos with a batch size of 4 at 256×256-pixel resolution, using 16 frames per video clip sampled at 4 fps with patch size of 16 and tubelet size of 2 for spatiotemporal tokenization. The model trains for 10 epochs with 300 iterations per epoch, totaling 3000 training iterations using mixed precision bfloat16 for computational efficiency. This pretraining approach enables the encoder to develop specialized representations for temporal dynamics through the robust keyframe sampling methodology.

　b) *LLM as text decoder Post-Training:*

Following the video encoder pretraining, an LLM decoder is fine-tuned to enable sports action description generation from the learned video representations. The approach adapts the framework from UI-JEPA by employing a frozen pretrained V-DWT-JEPA2 encoder combined with a fine-tuned language model decoder, creating a multimodal architecture capable of generating natural language descriptions of sports activities.

Combined with the 300M parameter video encoder and 3B parameter LLM, the total architecture comprises 3.3B parameters while maintaining efficiency through strategic parameter freezing and adaptation techniques. The vision projection consists of a two-layer multi-layer perceptron (MLP) that maps the 1024-dimensional video features through a 512-dimensional intermediate layer to the LLM's embedding dimension, incorporating ReLU activation, dropout regularization, and layer normalization for stable training. To maintain computational efficiency while enabling adaptation, LoRA (Low-Rank Adaptation) is applied to the LLaMA model with rank 16, targeting all linear layers including query, key, value, output, gate, up, and down projections. The fine-tuning employs a batch size of 8 with a learning rate of 1e-4, utilizing warmup steps and gradient clipping for stable optimization. This post-training phase enables the model to generate coherent sports action descriptions by leveraging the rich temporal representations learned during the self-supervised pretraining stage while maintaining computational efficiency.

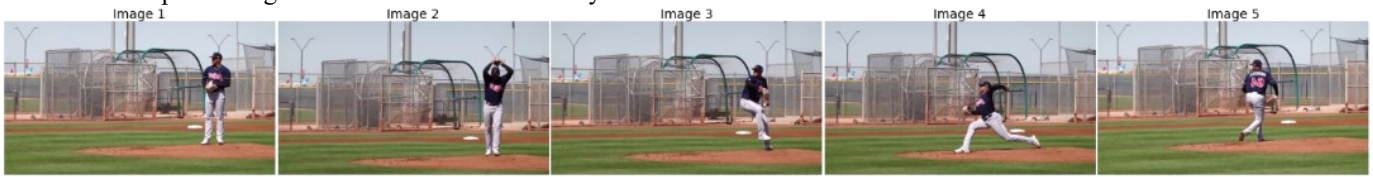

Fig. 3. Extracted key frames using DWT-VGG16-LDA.

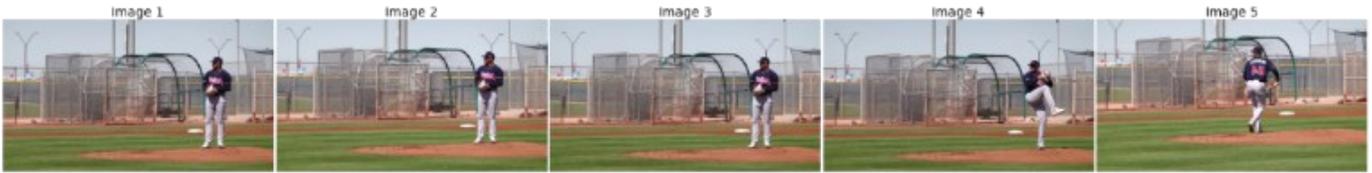

Fig. 4. Extracted key frames using uniform sampling.

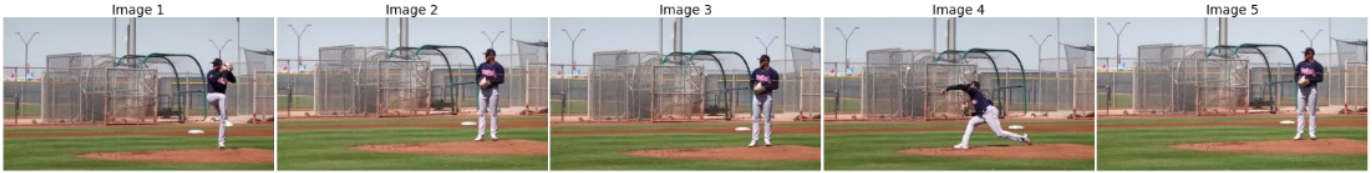

Fig. 5. Extracted key frames using color histogram.

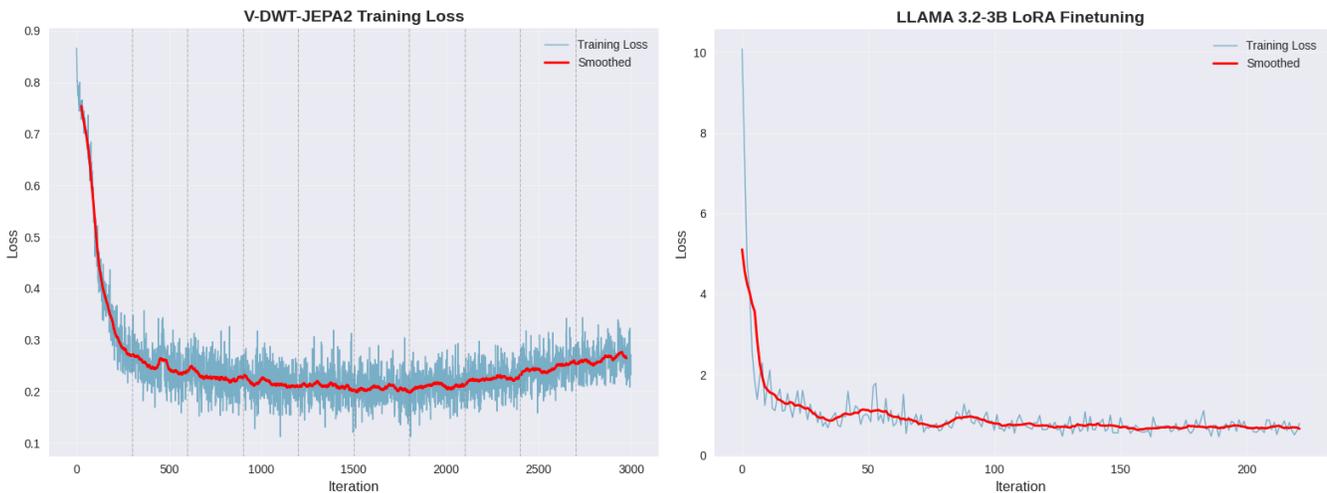

Fig. 6. Training loss curves for pre-training and post-training, showing raw losses (blue) and smoothed trends (red).

## IV. EMPIRICAL EVALUATION

### A. Dataset

We utilize a subset of the NSVA (NBA Sports Video Analysis) dataset [reference], which was originally collected by leveraging publicly available data from the NBA's official website. The original NSVA dataset employs a comprehensive web scraping approach to collect high-resolution (720P) video clips paired with detailed play-by-play descriptions from 132 games across 10 NBA teams during the 2018-2019 season, containing 32,019 video clips with associated action descriptions, player information, and contextual metadata.

For our SV3.3B training, we created a subset of NSVA [42] due to computational resource constraints, resulting in 1,315 video clips split into 1,050 training samples and 265 validation samples. To enhance model generalizability and focus on action understanding rather than player identification, we removed player names from the action annotations, retaining only the core basketball actions. The resulting annotations include precise action descriptions such as "MISS 3' Layup", "OFFENSIVE REBOUND", "MISS 16' Pullup Jump Shot", and "17' Jump Shot", enabling the model to learn fine-grained basketball action recognition without being biased toward specific players. Each video clip corresponds to a single basketball event or sequence of related events, providing temporally coherent training data essential for understanding multi-phase athletic movements.

### B. Training Dynamics

Fig. 6 illustrates the training loss progression for both phases of the V-DWT-JEPA2 model development. The left panel shows the self-supervised pretraining phase, where the encoder loss exhibits rapid initial convergence from approximately 0.8 to 0.25 within the first 500 iterations, followed by stable training with minor fluctuations around 0.25-0.3 for the remaining iterations. This convergence pattern indicates effective learning of spatiotemporal representations through the mask-denoising objective. The right panel demonstrates the LLM decoder fine-tuning phase, displaying characteristic behavior of language model adaptation with an initial sharp decrease from approximately 5.0 to 1.0 within the first 50 iterations, subsequently stabilizing around 0.5-1.0. The smoothed curves (red lines) reveal consistent downward trends in both phases, confirming stable training dynamics and successful convergence for both the self-supervised encoder pretraining and the supervised decoder fine-tuning components of the architecture.

*C. Evaluation Metrics*

The evaluation framework employs a comprehensive set of metrics designed to assess both the accuracy and richness of sports action descriptions generated by the model. These metrics are categorized into individual measures and composite scores that provide holistic performance assessment.

1) *Individual Metrics:*
   - *ROUGE-L F1:* Measures the longest common subsequence F1 score between predictions and ground truth, providing insight into structural similarity and word ordering accuracy commonly used in text summarization tasks.
   - *BLEU Score:* Evaluates n-gram precision with brevity penalty, particularly effective for assessing exact terminology matching crucial in sports commentary, following standard machine translation evaluation protocols.
   - *BERT F1:* Captures semantic similarity through contextual embeddings, enabling evaluation beyond surface-level text matching by leveraging pre-trained language model representations.
   - *Content F1:* Focuses on meaningful word overlap by excluding stop words, emphasizing sports-relevant vocabulary alignment between generated and reference descriptions.
   - *Semantic Similarity:* Employs cosine similarity of sentence embeddings to measure conceptual alignment, quantifying semantic coherence independent of lexical overlap.

2) *Sports-Specific Metrics:*
   - *Information Density:* Calculates the ratio of technical sports terms to total words, measuring the concentration of domain-specific vocabulary in generated descriptions.
   - *Action Complexity:* Counts distinct sports actions mentioned, assessing the model's ability to identify and describe multiple movement phases within athletic sequences.
   - *Measurement Precision:* Quantifies the use of quantitative details such as distances, times, and scores, reflecting the analytical depth and specificity of sports descriptions.
   - *Sequence Length:* Measures the number of distinct events described in chronological order, evaluating temporal understanding capabilities.
   - *Vocabulary Richness:* Evaluates diversity of sports terminology weighted by corpus frequency, indicating lexical sophistication in domain-specific language use.
   - *Technical Coverage:* Determines the percentage of available sports technical terms utilized from the total corpus vocabulary, measuring comprehensiveness of domain knowledge application.

3) *Composite Metrics:*
   - *Ground Truth Validation Score:* Combines ROUGE-L F1, BLEU Score, BERT F1, Content F1, and Semantic Similarity to assess overall accuracy against reference descriptions.
   - *Information Richness Score:* Aggregates Information Density, Action Complexity, Measurement Precision, Sequence Length, Vocabulary Richness, and Technical Coverage to evaluate analytical depth and technical sophistication.
   - *Combined Performance Score:* Represents the sum of Ground Truth Validation Score and Information Richness Score, providing unified assessment across accuracy and richness dimensions for comprehensive model comparison.

*D. Model Performance Analysis*

SV3.3B demonstrates strong performance across both traditional text generation metrics and sports-specific evaluation criteria on the NSVA basketball dataset subset. The model achieves a Ground Truth Validation Score of 2.124, indicating effective alignment with reference descriptions, and an Information Richness Score of 160.697, demonstrating superior capability in generating technically detailed sports content. The combined performance score of 162.821 reflects the model's balanced proficiency across accuracy and domain-specific analytical dimensions.

Analysis of individual metrics reveals notable strengths in semantic understanding, with a BERT F1 score of $0.856 \pm 0.052$ indicating robust contextual comprehension. The model excels in sports-specific criteria, achieving high Information Density ($0.978 \pm 0.053$) and Action Complexity ($2.307 \pm 1.161$), demonstrating effective utilization of technical terminology and identification of multiple movement phases. The Measurement Precision score of $2.557 \pm 0.808$ indicates the model's capability to incorporate quantitative details essential for comprehensive sports analysis. With an average response length of 45.2 words, SV3.3B generates concise yet analytically rich descriptions that capture the essential technical nuances of basketball actions while maintaining computational efficiency suitable for real-time deployment scenarios.

*E. Comparison to State-Of-The-Art*

To evaluate the model's performance, baseline comparisons were conducted using GPT-o4 Mini, GPT-4o Mini, and GPT-4o models with uniformly sampled video frames, while our SV3.3B model utilized the proposed DWT-based temporal sampling approach. The results were validated against the comprehensive metrics framework described above.

The findings as described in Tables I, II, III and IV demonstrates that SV3.3B achieves superior performance across both traditional text generation metrics and sports-specific evaluation criteria while maintaining significantly lower computational requirements compared to the larger closed-source baseline models.

TABLE I. GROUND TRUTH VALIDATION METRICS

| Model | ROUGE-L F1 | BLEU SCORE | BERT F1 | CONTENT F1 | SEMANTIC SIMILARITY | GT VALIDATION SCORE |
|---|---|---|---|---|---|---|
| GPT-o4 Mini | 0.212 ± 0.216 | 0.063 ± 0.116 | 0.732 ± 0.278 | 0.240 ± 0.249 | 0.467 ± 0.240 | 1.713 |
| GPT-4o Mini | 0.101 ± 0.108 | 0.016 ± 0.030 | 0.695 ± 0.287 | 0.112 ± 0.135 | 0.382 ± 0.198 | 1.305 |
| GPT-4o | 0.231 ± 0.239 | 0.033 ± 0.054 | 0.706 ± 0.299 | 0.223 ± 0.238 | 0.450 ± 0.263 | 1.643 |
| **SV3.3B (ours)** | **0.255 ± 0.248** | **0.103 ± 0.147** | **0.856 ± 0.052** | **0.350 ± 0.336** | **0.559 ± 0.235** | **2.123** |

TABLE II. INFORMATION RICHNESS METRICS

| Model | Information Density | Action Complexity | Measurement Precision | Sequence Length | Vocabulary Richness | Technical Coverage |
|---|---|---|---|---|---|---|
| GPT-o4 Mini | 0.816 ± 0.337 | 1.453 ± 1.189 | 0.146 ± 0.353 | 1.995 ± 1.546 | 113.8 ± 78.8 | 0.020 ± 0.014 |
| GPT-4o Mini | 0.654 ± 0.318 | 1.370 ± 1.214 | 1.188 ± 1.553 | **3.536 ± 1.655** | 80.9 ± 46.7 | **0.031 ± 0.015** |
| GPT-4o | 0.829 ± 0.356 | 1.651 ± 1.228 | 0.094 ± 0.325 | 1.745 ± 1.243 | 131.7 ± 90.9 | 0.017 ± 0.012 |
| **SV3.3B (ours)** | **0.978 ± 0.053** | **2.307 ± 1.161** | **2.557 ± 0.808** | 3.271 ± 0.456 | **151.6 ± 66.4** | 0.028 ± 0.010 |

TABLE III. COMPREHENSIVE SCORE ANALYSIS

| Rank | Model | GT Validation Score | Info Richness Score | Combined Score |
|---|---|---|---|---|
| 1 | **SV3.3B (ours)** | **2.1239** | **160.6970** | **162.8209** |
| 2 | GPT-o4 Mini | 1.6431 | 136.0328 | 137.6759 |
| 3 | GPT-4o Mini | 1.7134 | 118.2262 | 119.9397 |
| 4 | GPT-4o | 1.3056 | 87.6506 | 88.9562 |

TABLE IV. STATISTICAL ANALYSIS

| Comparison | GT Score Validation | Cohen's d | 95% CI Lower | 95% CI Upper |
|---|---|---|---|---|
| SV3.3B vs GPT-o4 Mini | +0.4105 | 1.24 | 0.31 | 0.51 |
| SV3.3B vs GPT-4o Mini | +0.8183 | 2.47 | 0.71 | 0.92 |
| SV3.3B vs GPT-4o | +0.4808 | 1.45 | 0.38 | 0.58 |

## V. CONCLUSION

In this work, we introduced SV3.3B, a novel framework that addresses the fundamental challenge of generating detailed, accurate sports action descriptions while operating under severe computational constraints suitable for edge device deployment. Our approach demonstrates that robust temporal sampling through DWT-based keyframe extraction, combined with self-supervised V-JEPA2 architecture and parameter-efficient fine-tuning, can achieve superior performance compared to computationally intensive closed-source models. The comprehensive evaluation across traditional text generation metrics and domain-specific sports analytics criteria confirms that SV3.3B successfully bridges the gap between professional-grade sports analytics and accessible technology. The model's ability to capture fine-grained biomechanical transitions and generate analytically rich descriptions while maintaining efficiency represents a significant advancement in democratizing sports video analysis. Future work will focus on extending the framework to additional sports domains, investigating longer temporal sequences, and exploring the integration of real-time feedback mechanisms for enhanced coaching applications. We believe SV3.3B establishes a new paradigm for lightweight multimodal sports understanding that can benefit both amateur and professional athletic development through accessible, high-quality video analysis capabilities.